\documentclass{article}
\usepackage[numbers]{natbib}
\usepackage[final]{nips_2018}
\usepackage[utf8]{inputenc} % allow utf-8 input
\usepackage[T1]{fontenc}    % use 8-bit T1 fonts
\usepackage{hyperref}       % hyperlinks
\usepackage{url}            % simple URL typesetting
\usepackage{booktabs}       % professional-quality tables
\usepackage{amsfonts}       % blackboard math symbols
\usepackage{nicefrac}       % compact symbols for 1/2, etc.
\usepackage{microtype}      % microtypography

\usepackage{amsmath}
\usepackage{algorithm}
\usepackage[noend]{algpseudocode}
\usepackage{epsfig}
\usepackage{graphicx}
\usepackage{subcaption}
\DeclareMathOperator{\argmax}{argmax}
\DeclareMathOperator{\argmin}{argmin}

\title{Dyna-AIL : Adversarial Imitation Learning by Planning}

\makeatletter
\newcommand{\printfnsymbol}[1]{%
  \textsuperscript{\@fnsymbol{#1}}%
}
\makeatother

\author{
  Vaibhav Saxena\thanks{equal contribution} \\
  MScAC, DCS\\
  University of Toronto\\
  \texttt{vaibhav@cs.toronto.edu} \\
  \And
  Srinivasan Sivanandan\printfnsymbol{1}\\
  MScAC, DCS\\
  University of Toronto\\
  \texttt{srinivasan@cs.toronto.edu} \\
  \And
  Pulkit Mathur\printfnsymbol{1}\\
  MScAC, DCS\\
  University of Toronto\\
  \texttt{pulkit@cs.toronto.edu} \\
}

\begin{document}

\maketitle

% \begin{abstract}
%   The abstract paragraph should be indented \nicefrac{1}{2}~inch
%   (3~picas) on both the left- and right-hand margins. Use 10~point
%   type, with a vertical spacing (leading) of 11~points.  The word
%   \textbf{Abstract} must be centered, bold, and in point size 12. Two
%   line spaces precede the abstract. The abstract must be limited to
%   one paragraph.
% \end{abstract}
\vspace{-1em}
\begin{abstract} %% TODO less

Adversarial methods for imitation learning have been shown to perform well on various control tasks. However, they require a large number of environment interactions for convergence. In this paper, we propose an end-to-end differentiable adversarial imitation learning algorithm in a Dyna-like framework for switching between model-based planning and model-free learning from expert data. Our results on both discrete and continuous environments show that our approach of using model-based planning along with model-free learning converges to an optimal policy with fewer number of environment interactions in comparison to the state-of-the-art learning methods.
\end{abstract}

\section{Introduction}
\label{introduction}
Reinforcement learning (RL) refers to the learning framework where an agent tries to learn the optimal policy by obtaining rewards from the environment through interaction. However, learning a policy from scratch is often difficult and involves numerous agent interactions with the environment. Further, in many cases, it is difficult to obtain explicit reward for actions taken from the environment. Imitation learning solves this problem by learning a policy from expert's state-action-state trajectories without access to reinforcement signal from the environment or interaction with the expert. Imitation is essential in applications including automation (e.g., imitating a human expert), distillation (e.g., if the expert is too expensive to run in real time \cite{RUSU2015}), and initialization (e.g., using an expert policy as an initial solution).

Reinforcement learning methods can be broadly classified into Model-Based (MB) methods involving planning with a world model and Model-Free (MF) methods involving learning by reactive execution in the environment. While MF methods can be used to learn complex policies in high-dimensional state spaces, they incur a huge cost in the number of interactions required with the environment. Use of a model reduces the number of such interaction trials drastically, but requires the dynamics model to be very accurate as a small bias in the model can lead to a strong bias in the policy.

Dyna \cite{Dyna-Sutton} based methods use a combined approach wherein the policy is learned by alternating between MB and MF methods. Using MB prior for MF RL (MBMF) is another kind of method that incorporates advantages of both MB and MF paradigms, by learning a probabilistic dynamics model which later acts as a prior for model-free optimization \cite{MBMF-GP}.

Ho and Ermon et al. (2016) proposed Generative Adversarial Imitation Learning (GAIL) \cite{GAIL}, an algorithm for performing imitation learning in a model-free setup using a GAN-like architecture, where the generator acts as the policy and the discriminator function judges whether an action given to it is from an expert or from the imitating policy. In this method, the agent learns the expert behaviour by an on-policy update on the state transition observations from the environment. However, in many practical learning situations, interactions with the environment are quite expensive. In this paper, we propose Dyna Adversarial Imitation Learning (Dyna-AIL), a framework for adversarial imitation learning by learning the expert policy by switching between world-model based planning and model-free reactive execution.

The remaining of this paper is organized as follows: in Section \ref{background}, we talk about the existing adversarial imitation learning algorithms, and in Section \ref{proposed_algorithm}, we introduce our proposed Dyna-AIL algorithm. Section \ref{experiments} gives the experiments that we performed to test our algorithm, and finally, in Section \ref{discussion}, we discuss the results obtained from our experiments alongside the strengths and limitations of our algorithm.

\section{Background}
\label{background}

\begin{figure}[t]
\centering
\includegraphics[width=0.50\textwidth]{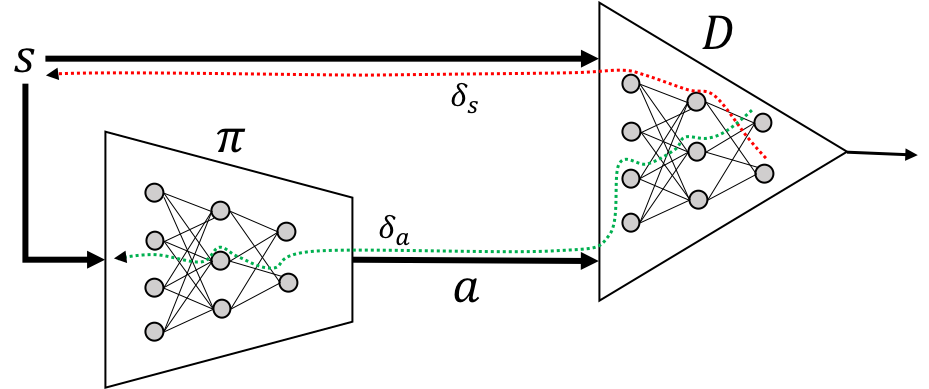}
\caption{Computation graph of GAIL showing how gradient w.r.t. state ($\delta_{s}$) is disregarded during back propagation.}
\label{fig:Gail}
\end{figure}

\subsection{Generative Adversarial Imitation Learning}
\label{GAIL}
Ho and Ermon (2016) proposed the Generative Adversarial Imitation Learning (GAIL) \cite{GAIL} architecture for learning policies from expert data using Generative Adversarial Networks \cite{GAN}. GAIL is a model-free approach wherein, the adversarial two-player zero-sum game can be represented as follows:

\begin{equation}
\label{GAIL obj}
\argmin_{\pi} \argmax_{D \in (0,1)} \mathbb{E}_{\pi}[\mathrm{log}(D(s,a))] + \mathbb{E}_{\pi_E}[\mathrm{log}(1-D(s,a))]
\end{equation}

A discriminator function $D$ tries to maximize the objective in (\ref{GAIL obj}), while a policy function $\pi$ tries to minimize it. In a neural network implementation, GAIL alternates between an Adam gradient \cite{kingma2014adam} step on the parameters of $D$, and a Trust Region Policy Optimization (TRPO) \cite{TRPO} gradient step on the parameters of $\pi$ which minimizes the cost $c(s,a) = \log D(s,a)$ for the policy. Fig. \ref{fig:Gail} shows how the gradients are passed from the discriminator down to the policy network. Since the state transitions are taken from the environment, the gradient w.r.t. state $s$ ($\delta_s$) does not pass through to the policy, or in other words, it is ignored and not used for the policy update.

\begin{figure}[t]
\centering
\includegraphics[width=0.9\textwidth]{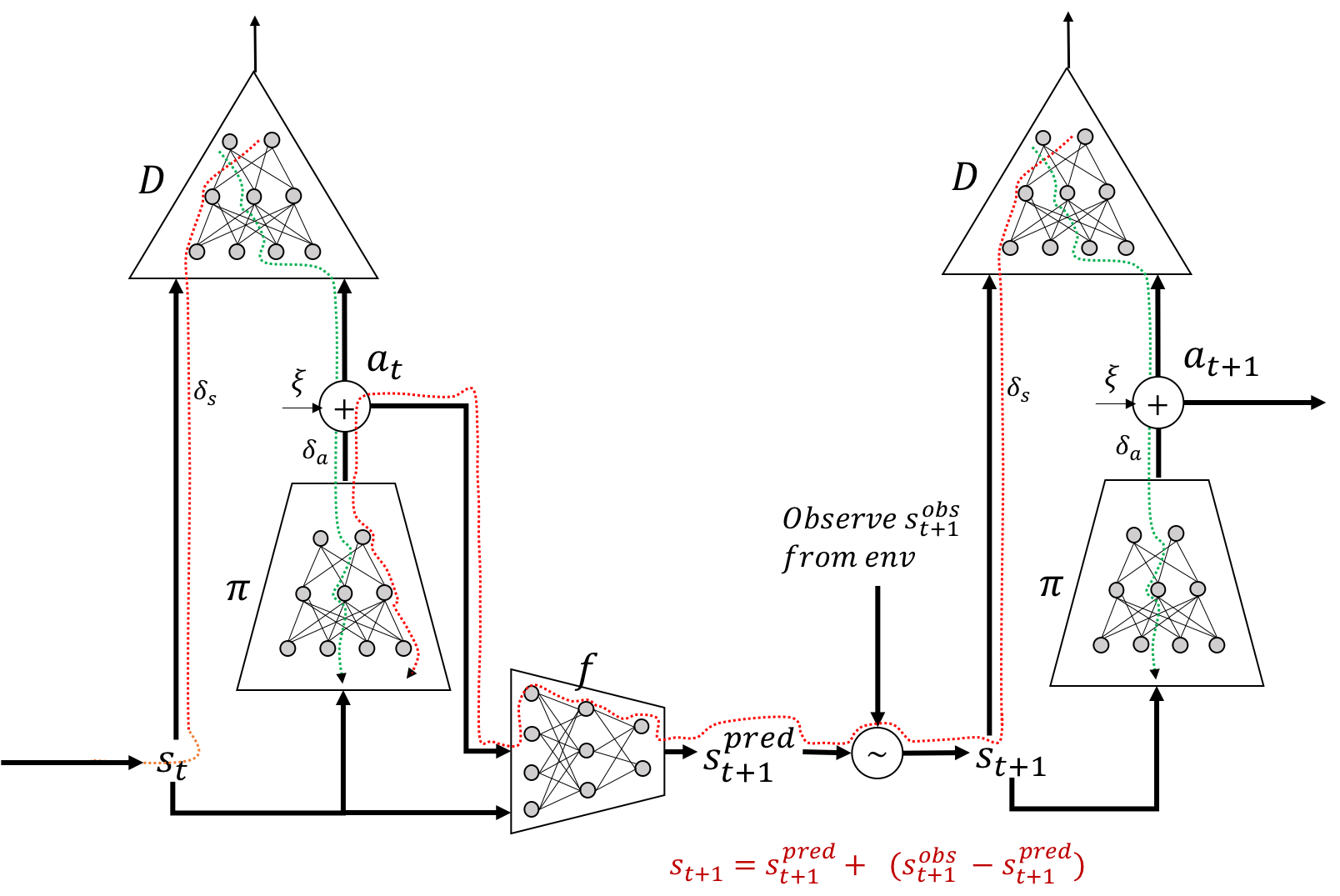}
\caption{Computation graph of the MGAIL algorithm showing the end-to-end gradient computation over multiple time steps with the re-paameterization trick}
\label{fig:MGail}
\end{figure}

\subsection{End-to-end differentiable Adversarial Imitation Learning (MGAIL)}
\label{MGAIL}

Baram et al. (2017) \cite{MGAIL} proposed an end-to-end differentiable version of GAIL where they use a forward-model through which the gradient of discriminator $D$ w.r.t. the state $s$ can be used for the policy update. They argue that since REINFORCE policy gradients suffer from high variance, it is difficult to work with them even after using variance reduction techniques. They also state that REINFORCE gradients are not same as the exact gradient of $D$, due to the independence assumption of state transitions w.r.t. policy parameters taken in REINFORCE. So, if we assume that the state transition probabilities are dependent on the policy parameters, i.e. use exact gradients through the incorporation of a forward-model, the policy will receive better signals while learning. They build an end-to-end differentiable computation graph that spans over multiple time-steps as given by Heess et al. (2015) \cite{SVGHeess} in their formulation, which for a transition ($s, a, s'$) is given as:

\begin{equation}
\label{eq: Heess d_objective_ds}
J_s=\mathbb{E}_{p(a|s)} \mathbb{E}_{p(s'|s,a)}
\Bigg[ D_s + D_a \pi_s + \gamma J'_{s'}(f_s+f_a\pi_s) \Bigg] ,
\end{equation}
\begin{equation}
\label{eq: Heess d_objective_dtheta}
J_\theta=\mathbb{E}_{p(a|s)} \mathbb{E}_{p(s'|s,a)}
\Bigg[ D_a\pi_\theta + \gamma (J'_{s'}f_a\pi_\theta + J'_\theta)\Bigg] .
\end{equation}

The gradient with respect to an entire trajectory is obtained by applying (\ref{eq: Heess d_objective_ds}) and (\ref{eq: Heess d_objective_dtheta}) recursively over the trajectory, and is illustrated in Fig. \ref{fig:MGail}. We note that MGAIL still uses state transitions from the environment while unrolling trajectories, probably because of an inadequacy in the forward-model capacity which would otherwise lead to noisy state transitions, and use a re-parameterization of environment state transitions so as to calculate gradients.

\begin{figure}[t]
\centering
\includegraphics[width=0.95\textwidth]{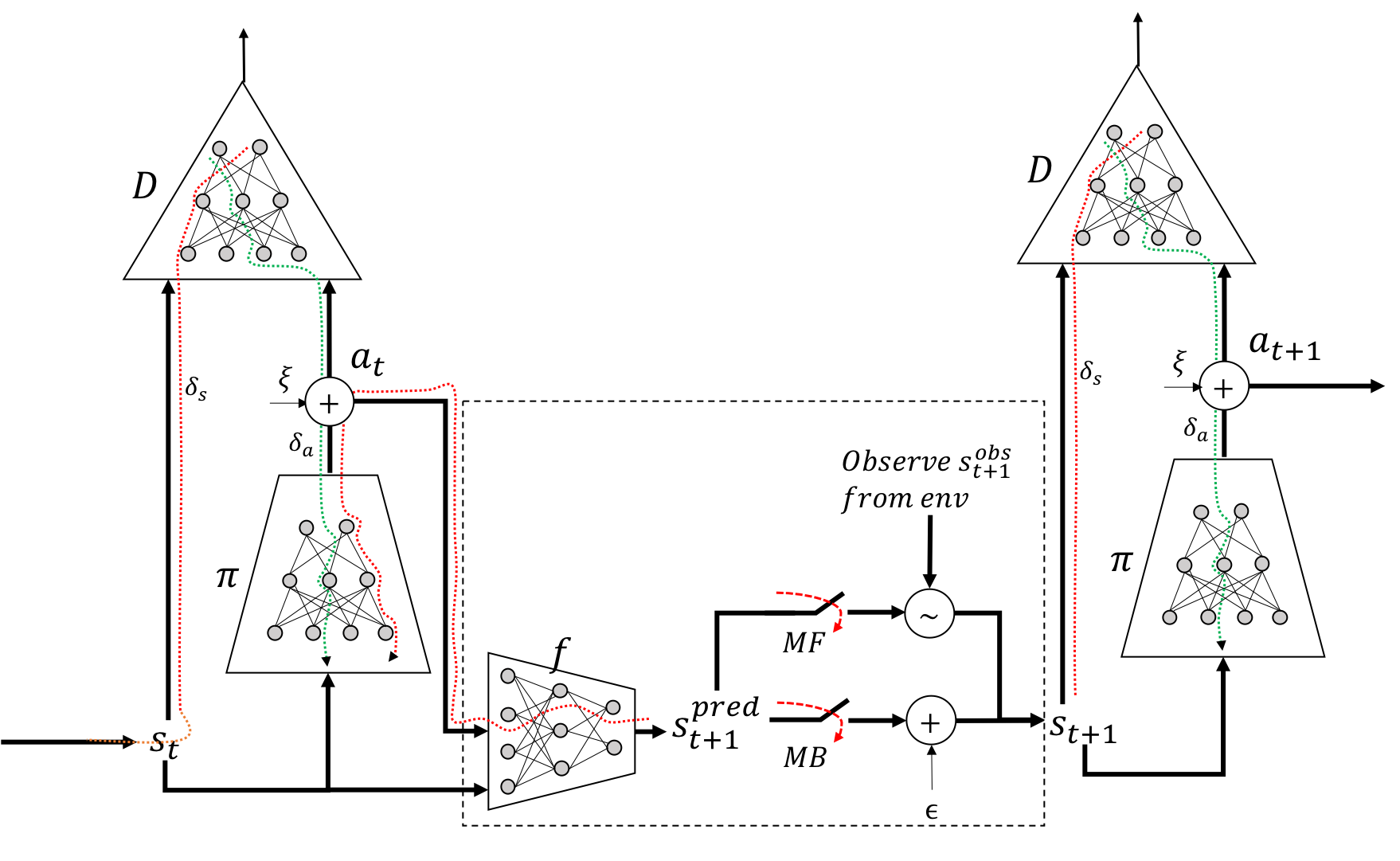}
\caption{Computation graph of our proposed Dyna-AIL method showing the switch between model-based (MB) planning and model-free (MF) learning by execution.}
\label{fig:DynAIL}
\end{figure}

\subsection{Comparing the Adversarial Imitation Learning algorithms}
\label{compare GAIL MGAIL}
Baram et al. (2017) added a forward-model in their algorithm so as to pass the gradient of discriminator $D$ w.r.t. state $s$ down to the policy network. However, they still sampled trajectories from the environment for computing the loss for the policy as well as the discriminator. In order to compute the gradients of the discriminator w.r.t. states, they re-parameterize the observed state $s^{obs}_{t+1}$ as $s^{pred}_{t+1} + \nu$, where $\nu = s^{obs}_{t+1} - s^{pred}_{t+1}$ as shown in Fig. \ref{fig:MGail}.
Hence, MGAIL, like GAIL, is a model-free algorithm w.r.t. the state transitions.

\section{Algorithm}
\label{proposed_algorithm}

The Dyna architecture proposed in \cite{Dyna-Sutton} integrates both model-based planning and model-free reactive execution to learn a policy. In this work, we present an algorithm (Algorithm \ref{Dyna-AIL}) for using the Dyna architecture with adversarial imitation learning methods to obtain improvement over environment sampling efficiency. Our aim is to learn a discriminator function $D$ and a policy function $\pi$ which respectively maximize and minimize the expression

\begin{equation}
\label{AIL obj}
    \mathbb{E}_{\pi}[\log(D(s,a))] + \mathbb{E}_{\pi_E}[\log(1-D(s,a))].
\end{equation}

We use a neural network function approximator for $D$ and $\pi$, and use a set of expert trajectories to calculate the expectation w.r.t. $\pi_E$. Now, in every iteration, we alternate between a learning phase and a planning phase. During the learning phase, we sample trajectories using $\pi$ on the \textit{environment} to make transitions, and take a gradient step on $D$ using (\ref{alg:D learning update}) in order to maximize (\ref{AIL obj}). Then we use an on-policy gradient update on $\pi$ using (\ref{alg:pi learning update}) so as to minimize (\ref{AIL obj}). In every iteration, we train the forward-model by minimizing the squared loss
\vspace{-0.15em}
\begin{equation}
\label{f squared loss}
    L_f = \sum_{(s,a,s') \in B} \frac{1}{2}||f(s,a) - (s' - s)||^2
\end{equation}
between the state transition predictions from the forward-model and the observed state transitions $(s, a, s')$ sampled from trajectories stored in the experience replay buffer $B$. To better represent the environment by a neural network function approximator, we designed the output of our forward-model as the change in the state values $f(s,a) = \delta{s_{pred}}$ such that $s_{pred}' = s + f(s,a)$.

\begin{algorithm}[H]
\caption{Dyna - Adversarial Imitation Learning}
\label{Dyna-AIL}
\begin{algorithmic}[1]
% \Procedure{MyProcedure}{}
\State \textbf{Input:} Expert trajectories $\tau_E$, experience buffer $B$, initial parameters for policy ($\pi$) and discriminator ($D$) $\theta_g, \theta_d$
% , num\_MF\_iters, num\_MB\_iters
\Repeat

%\For{i = 1 $\to$ num\_MF\_iters}
\State Sample trajectories $\tau_i$ using policy $\pi(a | s; \theta_g)$ on environment
\State Store trajectories $\tau_i$ into $B$ \Comment{for experience replay}
\State Update discriminator ($D$) parameters $\theta_d$ with gradient
\begin{equation}
\label{alg:D learning update}
    \hat{\mathbb{E}}_{\tau_i}[\nabla_{\theta_d} \log(D_{\theta_d}(s,a))] +             \hat{\mathbb{E}}_{\tau_E}[\nabla_{\theta_d} \log(1-D_{\theta_d}(s,a))]
\end{equation}
\State Update policy ($\pi$) parameters $\theta_g$ with gradient
\begin{equation}
\label{alg:pi learning update}
    \nabla_{\theta_g} \mathbb{E}_{\tau_i}\bigg[ \sum_{t=0} \gamma^t\log(D(s_t, a_t)) \bigg]
\end{equation}
%\EndFor

\State Train forward-model $f$ using $(s, a, s')$ from $B$
%% add loss fn

%\For{i = 1 $\to$ num\_MB\_iters}
\State Sample trajectories $\tau_j$ using policy $\pi(a | s; \theta_g)$ on forward model $f$
\State Update policy ($\pi$) parameters $\theta_g$ with gradient \Comment{model-based planning}
\begin{equation}
\label{alg:pi planning update}
    \nabla_{\theta_g} \mathbb{E}_{\tau_j}\bigg[\sum_{t=0} \gamma^t\log(D(s_t, a_t)) \bigg]
\end{equation}
%\EndFor

\Until{convergence}
\end{algorithmic}
\end{algorithm}

During the planning phase, we sample trajectories using $\pi$ on the \textit{forward-model} for state transitions, and take a gradient step on $\pi$ using (\ref{alg:pi planning update}). By having a planning step in every iteration, our approach optimizes the number of environment interactions required to imitate an expert policy, as shown in the experiments.

In our experiments, we use the multi-step computation graph with gradient updates as given in (\ref{eq: Heess d_objective_ds}) and (\ref{eq: Heess d_objective_dtheta}). However, we also discuss our algorithm's performance using natural policy gradient update with TRPO rule, in Section \ref{discussion}, for comparison with \cite{GAIL}.

\begin{figure}[t]
\begin{center}
    \begin{subfigure}[b]{0.5\textwidth}
    \includegraphics[width=\textwidth]{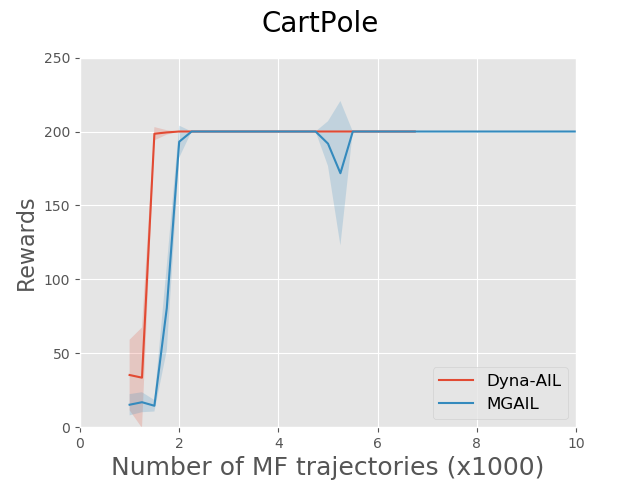}
    \caption{Cartpole environment.}
    \label{fig:cartpole dyna-ail vs mgail}
    \end{subfigure}
\end{center}

\begin{subfigure}[b]{0.5\textwidth}
    \includegraphics[width=\textwidth]{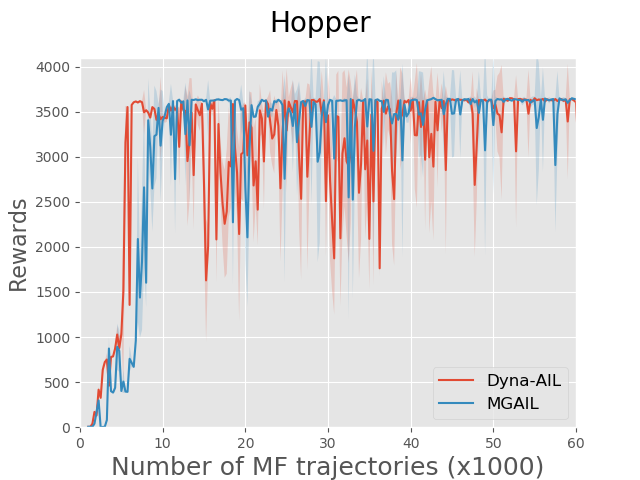}
    \caption{Hopper environment.}
    \label{fig:hopper dyna-ail vs mgail}
\end{subfigure}
\begin{subfigure}[b]{0.5\textwidth}
    \includegraphics[width=\textwidth]{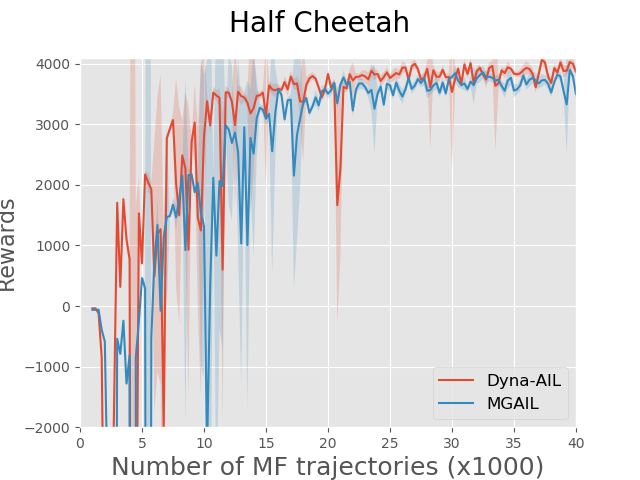}
    \caption{HalfCheetah environment.}
    \label{fig:half-cheetah dyna-ail vs mgail}
\end{subfigure}
\caption{Results of our experiments showing the rewards against number of model-free trajectories used while training Dyna-AIL vs MGAIL on discrete (cartpole) and continuous (Hopper, HalfCheetah) control tasks. (Trajectory length while planning $T_p = 10$)}
\label{fig:MGail comparisons}
\end{figure}

\section{Experiments and Results}
\label{experiments}
We evaluate our proposed algorithm on one discrete control task (CartPole \cite{barto1983neuronlike}), and two continuous control tasks (Hopper, HalfCheetah) modeled by the MuJoCo \cite{todorov2012mujoco} physics simulator, using the cost function defined in the OpenAI Gym \cite{gym}. We use the Trust Region Policy Optimization (TRPO) \cite{TRPO} algorithm to train our expert policies. For each of the tasks, we generated expert datasets with a number of trajectories ($n_T = 50$), where each trajectory: $\tau = {s_0, a_0, s_1, ...s_N , a_N }$ is of length N = 1000. 

The discriminator and policy networks are each designed with two hidden layers (200 and 100 units for discriminator, and 100 and 50 units for policy) with ReLU non-linearity and trained using the Adam \cite{kingma2014adam} optimizer (similar to the architectures used in \cite{MGAIL}). As noted in \cite{MGAIL}, the forward-model structure is crucial for the stability of the network's learning process. Since the state and action inputs are from different distributions, we embed them into a shared space using state and action encoder networks. Then, we combine the embeddings through a Hadamard product and use that as input for the state transition model. We also model the environment as a n$^{\mathrm{th}}$ order MDP, with a recurrent connection from the previous states, using a GRU layer as a part of the state encoder (as previously used in \cite{MGAIL}).

In every iteration of Algorithm \ref{Dyna-AIL}, we perform the updates in (\ref{alg:D learning update}) and (\ref{alg:pi learning update}) using a $m$-step ($m = 50$) stochastic gradient descent with trajectories sampled from the environment. Later in the iteration, we perform planning in (\ref{alg:pi planning update}) using a $p$-step ($p = 50$) stochastic gradient descent with trajectories sampled from the forward-model. To ensure the stability of the planner, we restrict the unrolling of the trajectory in planning phase to a fixed number of steps ($T_p$). In Section \ref{discussion}, we discuss the performance of our algorithm by varying the trajectory lengths $T_p$.

After every iteration, the policy learned in the algorithm was evaluated using 10 episodes by acting on the environment. Fig. \ref{fig:MGail comparisons} shows the learning curve comparison between our algorithm and MGAIL. From the experimental results, we observe that our algorithm learns the optimal policy with fewer number of interactions with the environment, as compared to MGAIL. However, in the case of Hopper environment, our algorithm has a high variance in performance, which could be attributed to the bias introduced by the forward-model used for sampling trajectories in the planning phase. In Section \ref{discussion}, we discuss a few possible solutions for overcoming the instability issues.

\begin{figure}[t]
% \centering
\begin{subfigure}[b]{0.5\textwidth}
    \includegraphics[width=\textwidth]{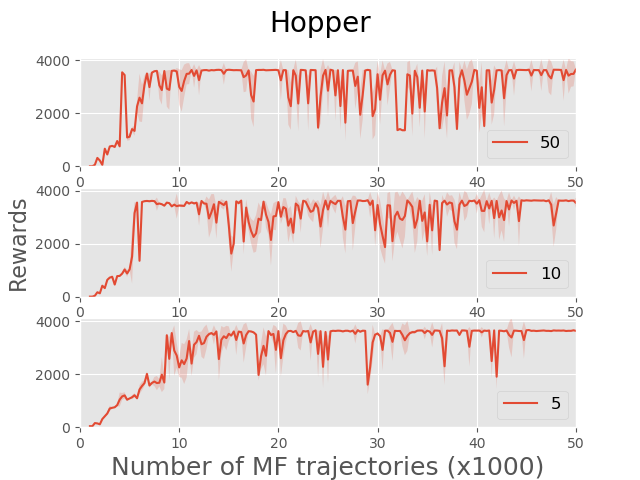}
    \caption{Hopper environment.}
    \label{fig:hopper dyna-ail traj lengths}
\end{subfigure}
\begin{subfigure}[b]{0.5\textwidth}
    \includegraphics[width=\textwidth]{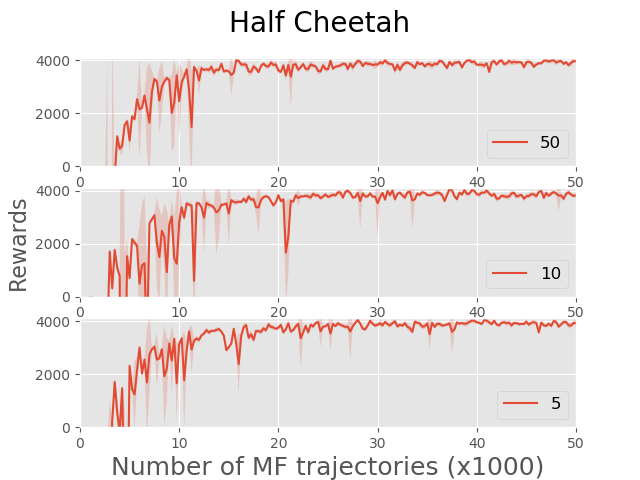}
    \caption{HalfCheetah environment.}
    \label{fig:half-cheetah dyna-ail traj lengths}
\end{subfigure}
\caption{Results of our experiments showing the rewards against number of model-free trajectories used while training Dyna-AIL, for different lengths of trajectories ($T_p$) sampled from the forward-model.}
\label{fig:dyna-ail traj length comparisons}
\end{figure}

\begin{figure}[t]
\centering
\includegraphics[width=0.62\textwidth]{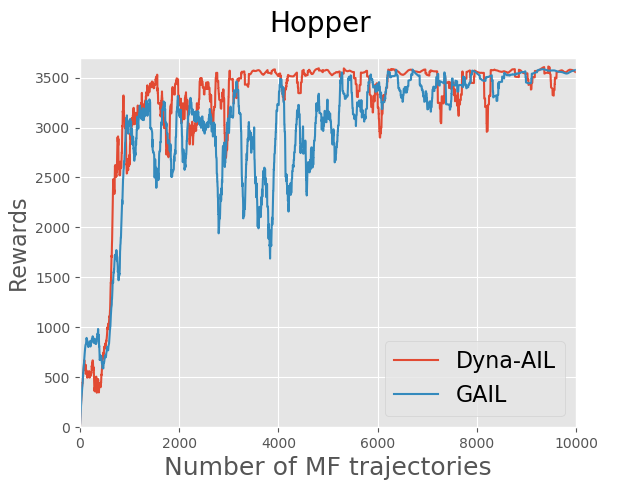}
\caption{Results of our experiments showing the rewards against number of model-free trajectories used while training Dyna-AIL vs GAIL (using TRPO gradient step) on Hopper task.}
\label{fig:Gail comparisons}
\end{figure}

\section{Discussion}
\label{discussion}
To evaluate the stability of our Dyna planner w.r.t the planning phase trajectory length $T_p$, we performed experiments with different values of $T_p$ on Hopper and HalfCheetah environments. Fig. \ref{fig:dyna-ail traj length comparisons} shows the learning curves for $T_p = 5, 10, 50$. We observe that with the Hopper environment, the learning curve has a higher variance with $T_p = 50$ as compared to with lower values of $T_p$. However, with the HalfCheetah environment, learning curves for all $T_p$ values converge with very low variance. We attribute this behavior to the capacity of the forward-model, which learns a good approximation for the HalfCheetah environment, but is unable to learn the same for the more complex Hopper environment which has a larger state-space. We intend to resolve this issue in future work by engineering the capacity of the forward-model and training it efficiently.

Even though MGAIL addresses the problem of high variance gradients by utilizing the forward-model gradients, they still use a small step size due to the gradient update as per \cite{SVGHeess}. GAIL copes with the high variance in gradients by performing the policy update using a natural policy gradient step with TRPO update rule. The KL divergence constraint enables us to use a large step size for policy update without worrying about the noisy policy gradients. Hence, we evaluated Dyna-AIL on Hopper environment without forward-model gradients using the TRPO update rule, as per \cite{GAIL}. Fig. \ref{fig:Gail comparisons} shows the performance of the Dyna-AIL algorithm with TRPO update compared with the results of GAIL. The TRPO update ensures that the policy updates do not diverge by adjusting the step size in every gradient update and hence reduces the variance in the Hopper environment. Further, in our future work, we intend to implement an end-to-end differentiable multiple-step computation graph with a natural policy gradient update as per the TRPO rule. 

In this paper, we proposed a framework to train policies via imitation learning by switching between model-based planning and model-free execution. The results show that this framework reduces the number of environment interactions required to learn a good policy while imitating an expert. However, we still perform only one planning step for every learning step as there is no direct metric to evaluate the quality of the planning step to stop planning and learn from the environment. Yuexin et. al. (2018) \cite{adapiveDynaMicrosoft}, propose an adaptive Dyna-Q framework by integrating a switcher that automatically determines whether to use a real or simulated experience for Q-learning based on a quality metric. A possible future direction is to have an adaptive switching mechanism in our algorithm that could query the environment only when the model is uncertain about a state transition.
\vspace{-0.5em}
% \section*{References}
\small
\bibliographystyle{unsrtnat}%ieeetr
\bibliography{DynaAIL}

\begin{thebibliography}{13}
\providecommand{\natexlab}[1]{#1}
\providecommand{\url}[1]{\texttt{#1}}
\expandafter\ifx\csname urlstyle\endcsname\relax
  \providecommand{\doi}[1]{doi: #1}\else
  \providecommand{\doi}{doi: \begingroup \urlstyle{rm}\Url}\fi

\bibitem[Rusu et~al.(2015)Rusu, Colmenarejo, G{\"{u}}l{\c{c}}ehre, Desjardins,
  Kirkpatrick, Pascanu, Mnih, Kavukcuoglu, and Hadsell]{RUSU2015}
Andrei~A. Rusu, Sergio~Gomez Colmenarejo, {\c{C}}aglar G{\"{u}}l{\c{c}}ehre,
  Guillaume Desjardins, James Kirkpatrick, Razvan Pascanu, Volodymyr Mnih,
  Koray Kavukcuoglu, and Raia Hadsell.
\newblock Policy distillation.
\newblock \emph{CoRR}, abs/1511.06295, 2015.
\newblock URL \url{http://arxiv.org/abs/1511.06295}.

\bibitem[Sutton(1991)]{Dyna-Sutton}
Richard~S. Sutton.
\newblock Dyna, an integrated architecture for learning, planning, and
  reacting.
\newblock \emph{SIGART Bull.}, 2\penalty0 (4):\penalty0 160--163, July 1991.
\newblock ISSN 0163-5719.
\newblock \doi{10.1145/122344.122377}.
\newblock URL \url{http://doi.acm.org/10.1145/122344.122377}.

\bibitem[Bansal et~al.(2017)]{MBMF-GP}
Somil Bansal et~al.
\newblock {MBMF:} model-based priors for model-free reinforcement learning.
\newblock \emph{CoRR}, abs/1709.03153, 2017.
\newblock URL \url{http://arxiv.org/abs/1709.03153}.

\bibitem[Ho and Ermon(2016)]{GAIL}
Jonathan Ho and Stefano Ermon.
\newblock Generative adversarial imitation learning.
\newblock \emph{CoRR}, abs/1606.03476, 2016.
\newblock URL \url{http://arxiv.org/abs/1606.03476}.

\bibitem[Goodfellow et~al.(2014)Goodfellow, Pouget-Abadie, Mirza, Xu,
  Warde-Farley, Ozair, Courville, and Bengio]{GAN}
Ian Goodfellow, Jean Pouget-Abadie, Mehdi Mirza, Bing Xu, David Warde-Farley,
  Sherjil Ozair, Aaron Courville, and Yoshua Bengio.
\newblock Generative adversarial nets.
\newblock In \emph{Advances in neural information processing systems}, pages
  2672--2680, 2014.

\bibitem[Kingma and Ba(2014)]{kingma2014adam}
Diederik~P. Kingma and Jimmy Ba.
\newblock Adam: {A} method for stochastic optimization.
\newblock \emph{CoRR}, abs/1412.6980, 2014.
\newblock URL \url{http://arxiv.org/abs/1412.6980}.

\bibitem[Schulman et~al.(2015)Schulman, Levine, Moritz, Jordan, and
  Abbeel]{TRPO}
John Schulman, Sergey Levine, Philipp Moritz, Michael~I. Jordan, and Pieter
  Abbeel.
\newblock Trust region policy optimization.
\newblock \emph{CoRR}, abs/1502.05477, 2015.
\newblock URL \url{http://arxiv.org/abs/1502.05477}.

\bibitem[Baram et~al.(2017)]{MGAIL}
Nir Baram et~al.
\newblock End-to-end differentiable adversarial imitation learning.
\newblock In \emph{Proceedings of the 34th ICML}, volume~70, pages 390--399.
  PMLR, 06--11 Aug 2017.
\newblock URL \url{http://proceedings.mlr.press/v70/baram17a.html}.

\bibitem[Heess et~al.(2015)Heess, Wayne, Silver, Lillicrap, Tassa, and
  Erez]{SVGHeess}
Nicolas Heess, Greg Wayne, David Silver, Timothy~P. Lillicrap, Yuval Tassa, and
  Tom Erez.
\newblock Learning continuous control policies by stochastic value gradients.
\newblock \emph{CoRR}, abs/1510.09142, 2015.
\newblock URL \url{http://arxiv.org/abs/1510.09142}.

\bibitem[Barto et~al.(1983)Barto, Sutton, and Anderson]{barto1983neuronlike}
Andrew~G Barto, Richard~S Sutton, and Charles~W Anderson.
\newblock Neuronlike adaptive elements that can solve difficult learning
  control problems.
\newblock \emph{Systems, Man and Cybernetics, IEEE Transactions on}, \penalty0
  (5):\penalty0 834--846, 1983.

\bibitem[Todorov et~al.(2012)Todorov, Erez, and Tassa]{todorov2012mujoco}
Emanuel Todorov, Tom Erez, and Yuval Tassa.
\newblock Mujoco: A physics engine for model-based control.
\newblock \emph{2012 IEEE/RSJ International Conference on Intelligent Robots
  and Systems}, pages 5026--5033, 2012.

\bibitem[Brockman et~al.(2016)Brockman, Cheung, Pettersson, Schneider,
  Schulman, Tang, and Zaremba]{gym}
Greg Brockman, Vicki Cheung, Ludwig Pettersson, Jonas Schneider, John Schulman,
  Jie Tang, and Wojciech Zaremba.
\newblock Openai gym.
\newblock \emph{CoRR}, abs/1606.01540, 2016.
\newblock URL \url{http://arxiv.org/abs/1606.01540}.

\bibitem[Wu et~al.(2018)Wu, Li, Liu, Gao, and Yang]{adapiveDynaMicrosoft}
Yuexin Wu, Xiujun Li, Jingjing Liu, Jianfeng Gao, and Yiming Yang.
\newblock Switch-based active deep dyna-q: Efficient adaptive planning for
  task-completion dialogue policy learning.
\newblock \emph{CoRR}, abs/1811.07550, 2018.
\newblock URL \url{http://arxiv.org/abs/1811.07550}.

\end{thebibliography}

% [1] Alexander, J.A.\ \& Mozer, M.C.\ (1995) Template-based algorithms
% for connectionist rule extraction. In G.\ Tesauro, D.S.\ Touretzky and
% T.K.\ Leen (eds.), {\it Advances in Neural Information Processing
%   Systems 7}, pp.\ 609--616. Cambridge, MA: MIT Press.

% [2] Bower, J.M.\ \& Beeman, D.\ (1995) {\it The Book of GENESIS:
%   Exploring Realistic Neural Models with the GEneral NEural SImulation
%   System.}  New York: TELOS/Springer--Verlag.

% [3] Hasselmo, M.E., Schnell, E.\ \& Barkai, E.\ (1995) Dynamics of
% learning and recall at excitatory recurrent synapses and cholinergic
% modulation in rat hippocampal region CA3. {\it Journal of
%   Neuroscience} {\bf 15}(7):5249-5262.

\end{document}